\pdfoutput=1

\documentclass[11pt]{article}
\IfFileExists{emnlp2023.sty}{%
  \usepackage{emnlp2023} 
}{%
  \usepackage{natbib}
}

\usepackage{times}
\usepackage{latexsym}
\usepackage[T1]{fontenc}
\usepackage[utf8]{inputenc}
\usepackage{microtype}
\usepackage{booktabs}
\usepackage{multirow}
\usepackage{amsmath}
\usepackage{graphicx}
\usepackage{tikz}
\usetikzlibrary{shapes.geometric, arrows.meta, positioning, calc, fit, backgrounds}

\title{MIPIAD: Multilingual Indirect Prompt Injection Attack Defense\\
with Qwen--TF-IDF Hybrid and Meta-Ensemble Learning}

\author{Al Muhit Muhtadi$^1$ and Mostafa Rifat Tazwar$^2$ \\
\small $^{1,2}$Bangladesh University of Engineering and Technology}

\begin{document}
\maketitle

\begin{abstract}
Indirect prompt injection remains a persistent weakness in retrieval-augmented and tool-using LLM systems, and the problem becomes harder to characterise in multilingual settings.
We present \textbf{MIPIAD}, a defense framework evaluated on English and Bangla that combines a sequence classifier fine-tuned from \texttt{Qwen2.5-1.5B} via LoRA (XLPID), TF-IDF lexical features, and validation-tuned ensembling through late fusion, stacking, and gradient boosting.
The framework is evaluated on a synthetic benchmark built from BIPIA \citep{yi2023benchmarking} templates spanning five task families---email, table, QA, abstract, and code---comprising over 1.43 million generated samples, with train and test splits using mutually exclusive attack categories.
Across the experiments, lexical signals prove unexpectedly strong (TF-IDF+SVM F1=0.77), and the hybrid XLPID+TF-IDF ensemble achieves the best overall F1 (0.9205) while the Boosting Ensemble achieves the best AUROC (0.9378).
Ensemble methods consistently reduce the English--Bangla cross-lingual gap relative to standalone neural models.
The pipeline is designed for extensibility: NLLB-200 supports over 200 languages and XLPID's multilingual backbone can be retargeted to additional languages without architectural changes; empirical validation is currently limited to English and Bangla.
\end{abstract}

\section{Introduction}
Prompt injection can make an LLM ignore the user's intent, expose sensitive information, or carry out unsafe tool actions.
The harder case is \emph{indirect} prompt injection, where malicious instructions are hidden inside retrieved documents, emails, tables, web pages, or code rather than in the user's query \citep{perez2022ignore}.

That setting is still underexplored once the input stream is multilingual.
In practice, systems see a mix of languages, domains, and formatting styles, so a defense that works only in one language or one surface is not enough.
This paper studies that more realistic setting and presents \textbf{MIPIAD} (Multilingual Indirect Prompt Injection Attack Defense), a full pipeline for cross-lingual detection and evaluation.
MIPIAD is empirically validated on English and Bangla; the underlying components---NLLB-200 translation and a multilingual LLM backbone---are designed to extend to additional languages without architectural modification.

We make four contributions:
\begin{itemize}
    \item A unified multilingual data pipeline that generates English/Bangla indirect prompt-injection samples across five tasks, forming a massive dataset of over 1.43M templates.
    \item \textbf{XLPID}, a cross-lingual prompt injection detector utilizing efficient Low-Rank Adaptation (LoRA) over robust LLM backbones.
    \item Meta-ensemble strategies (stacking and boosting) integrating LLM-based probability metrics with explicit TF-IDF lexical priors.
    \item Strong baseline comparisons showing that hybridization outperforms isolated neural models (F1: 0.9205 vs.\ 0.8939 for the best standalone, a 2.7-point gain).
\end{itemize}

\section{Related Work}
\textbf{Indirect Prompt Injection.}
Early work established prompt injection as a major security threat \citep{perez2022ignore}, which quickly proved highly applicable to real-world Retrieval-Augmented Generation (RAG) and LLM-integrated agent applications via indirect injections \citep{greshake2023indirect}.
To standardize evaluation, researchers introduced tailored benchmarks such as LLM-PIRATE \citep{ramakrishna2024llmpirate} for comprehensive evaluations and LLMail-Inject \citep{abdelnabi2025llmail} for realistic, adaptive threat scenarios.

\textbf{Defensive Mechanisms.}
Defenses fall into three families.
\emph{Input-filtering}: PromptGuard \citep{meta2024promptguard} screens inputs with a fine-tuned BERT classifier; SpotLight \citep{spotlightdefense2023} injects formatting markers into trusted context to help the LLM distinguish it from retrieved content.
\emph{Instruction-isolation}: InstructDetector \citep{wen2025instructdetector} identifies instruction states to limit injection blast radius; CachePrune \citep{wang2025cacheprune} prunes malicious context via neuron-level attribution.
\emph{Output-smoothing}: SmoothLLM \citep{robey2023smoothllm} applies input perturbation with majority voting, though at high inference cost and without evaluation on indirect retrieval injection; MELON \citep{zhu2025melon} protects agentic tool-use trajectories.
MIPIAD belongs to the input-filtering family, but uniquely combines lexical and neural signals, spans five task families, and operates on multilingual inputs.
None of the above defenses has been evaluated on Bangla or on large-scale multilingual indirect injection benchmarks; Section~\ref{sec:victim-eval} provides the first cross-lingual end-to-end victim evaluation for this class of defense.

\textbf{Multilingual Safety Gap.}
Most existing literature still assumes an English-heavy setting. Cross-lingual transfer is less understood for prompt-injection defense, where translations and domain shifts can dilute injection signatures and increase the false negative rate.
To our knowledge, prior work has not united the \emph{multilingual} and \emph{indirect} axes into a generalized detection framework.

\section{MIPIAD Benchmark}
\subsection{Threat Model and Task Definition}
We study binary detection over text inputs where the label $y \in \{0,1\}$ indicates whether an indirect injection is present.
Each sample is language-specific (EN or BN) and includes metadata for task, attack type, and insertion position. The defender's goal is to accurately detect embedded instructions before downstream victim LLMs consume them.

\begin{figure*}[t]
    \resizebox{\textwidth}{!}{
    \begin{tikzpicture}[
        node distance=1.3cm and 1.8cm,
        every node/.style={font=\small},
        block/.style={rectangle, rounded corners, draw=black!70, fill=blue!10, text width=3.5cm, align=center, minimum height=1cm},
        arrow/.style={-{Stealth}, thick}
    ]
        \node[block] (attacks) {English Attack Templates\\(BIPIA \citep{yi2023benchmarking})};
        \node[block, right=of attacks] (translateA) {NLLB-200 Translation\\(EN to BN)};
        \node[block, right=of translateA] (bilingA) {Bilingual Attack Texts};
        
        \node[block, below=0.5cm of attacks] (contexts) {Task Contexts (EN)\\(Email, Tables, QA, Abstract, Code)};
        \node[block, right=of contexts] (translateC) {NLLB-200 Translation\\(EN to BN)};
        \node[block, right=of translateC] (bilingC) {Bilingual Contexts};
        
        \node[block, right=1.0cm of bilingA, text width=3.8cm, fill=green!10] (compose) {Sample Composition\\(Inject at Start/Mid/End)};
        \node[block, right=1.0cm of bilingC, text width=3.8cm, fill=purple!10] (dataset) {MIPIAD Dataset\\(1.43M Raw Samples)};
        
        \draw[arrow] (bilingA.east) -- (compose.west);
        \draw[arrow] (bilingC.east) -- (compose.west);
        \draw[arrow] (compose.south) -- (dataset.north);
        
        \draw[arrow] (attacks) -- (translateA);
        \draw[arrow] (translateA) -- (bilingA);
        \draw[arrow] (contexts) -- (translateC);
        \draw[arrow] (translateC) -- (bilingC);
    \end{tikzpicture}
    }
    \caption{MIPIAD data generation pipeline. English strings and contexts are translated to Bangla using NLLB-200, then contextually composed to create a balanced bilingual dataset.}
\label{fig:pipeline}
\end{figure*}
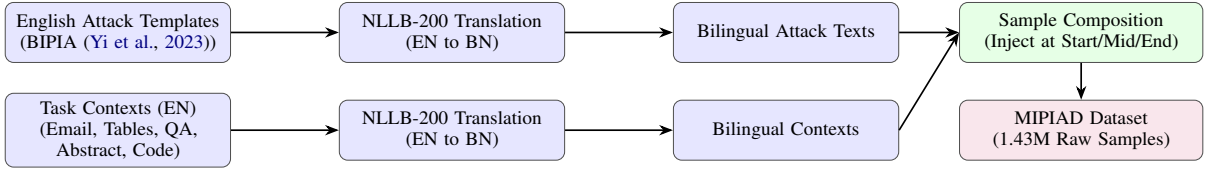

\subsection{Multilingual Sample Construction}
Figure~\ref{fig:pipeline} illustrates the data engineering pipeline. MIPIAD generation follows three main steps:
\begin{enumerate}
    \item Translate attack templates from English to Bangla with Meta's NLLB-200 \citep{nllb2022}.
    \item Translate task contexts across five families (email, table, QA, abstract, code) into Bangla.
    \item Compose poisoned samples by injecting attack texts at the start, middle, or end of the contexts; generate benign samples from clean contexts.
\end{enumerate}

To ensure robustness, our data generator yields an extensive matrix: 15 unique text attack categories and 10 code-specific attack categories, each with 5 variants. Combining these with 3 insertion positions and 2 languages (EN/BN) results in exactly 1,431,400 raw samples.

Crucially, to prevent cross-lingual data leakage, train and test splits are explicitly partitioned at the context level. All language translations for a single context are securely bounded within identical splits. Furthermore, to evaluate genuine generalization capabilities rather than rote memorization, the training and testing sets utilize completely mutually exclusive attack categories and variants.

\section{Methodology}

\subsection{XLPID Architecture}
\begin{figure}[t]
    \resizebox{\columnwidth}{!}{
    \begin{tikzpicture}[
        node distance=0.4cm,
        every node/.style={font=\small},
        block/.style={rectangle, rounded corners=4pt, draw=black!70, fill=blue!6, minimum width=15em, minimum height=2em, text centered},
        ioblock/.style={rectangle, rounded corners=4pt, draw=black!70, fill=green!8, minimum width=15em, minimum height=1.8em, text centered},
        arr/.style={-{Stealth[length=5pt]}, thick, draw=black!55},
    ]
        \node[ioblock] (input) {Input Document (EN/BN)};
        \node[block, below=of input, minimum height=2.4em] (encoder) {LLM Backbone (e.g., Qwen)\\{\scriptsize \textit{Weights frozen in bf16}}};
        \node[block, right=0.3cm of encoder, fill=orange!15, text width=2cm, minimum width=2.2cm, minimum height=2.4em] (lora) {LoRA\\{\scriptsize \textit{fp32}}};
        \node[block, below=of encoder] (pool) {Context Pooling};
        \node[block, below=of pool] (cls) {Sequence Classification Head};
        \node[ioblock, below=of cls, fill=purple!10] (out) {Injection Logits ($p_t$)};

        \draw[arr] (input) -- (encoder);
        \draw[arr] (encoder) -- (pool);
        \draw[arr] (pool) -- (cls);
        \draw[arr] (cls) -- (out);
        \draw[dashed, thick, Stealth-Stealth, draw=black!55] (encoder) -- (lora);
    \end{tikzpicture}
    }
    \caption{XLPID architecture utilizing parameter-efficient LoRA adapters alongside a sequential classification head.}
\label{fig:xlpid-arch}
\end{figure}
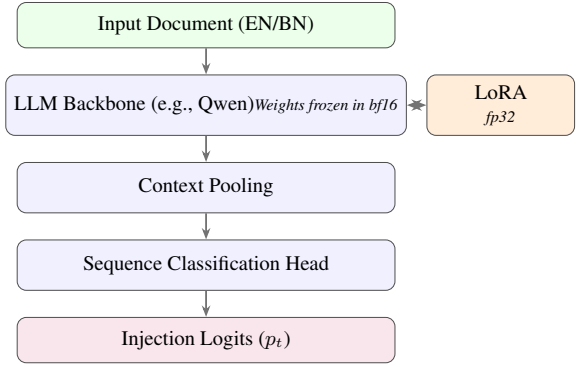

The core neural detector of our framework is \textbf{XLPID} (Cross-Lingual Prompt Injection Detector). Figure~\ref{fig:xlpid-arch} provides a structural overview of the classification scheme.

XLPID is a direct sequence-classification wrapper over a frozen LLM backbone, using the backbone's built-in classification head (context pooling layer followed by a two-label linear projection). XLPID supports multiple backbone families including Qwen2.5 \citep{qwen2025} and DeBERTa \citep{he2021deberta}; all results in this paper use \textbf{Qwen/Qwen2.5-1.5B} as the backbone. Base weights are kept in \texttt{bfloat16} to reduce VRAM while LoRA adapters (rank=16, $\alpha$=32) targeting \texttt{q\_proj} and \texttt{v\_proj} are trained in \texttt{float32}.

\subsection{Meta-Ensembles and Lexical Baselines}
We compare XLPID against isolated contextual backbones (XLM-RoBERTa \citep{liu2019roberta}, mBERT \citep{devlin2019bert}) and powerful lexical baselines (TF-IDF + LR, TF-IDF + SVM) comprising 10,000 top n-gram features (sizes 1--3).

Furthermore, we evaluate two meta-ensembles synthesizing transformer and lexical streams into a unified prediction:
\begin{itemize}
    \item \textbf{Hybrid late fusion}: Combining XLPID transformer probabilities ($p_t$) and TF-IDF probabilities ($p_l$) via $p = \alpha p_t + (1-\alpha)p_l$. The mixing weight $\alpha$ is selected by grid search over 21 evenly spaced values in $[0,1]$ (i.e.\ $\alpha \in \{0.00, 0.05, \ldots, 1.00\}$), evaluated on the held-out validation set (10\% of training data) and maximising the composite criterion $(\text{F1}, \text{AUROC})$ lexicographically. The best $\alpha$ is then locked in before any test-set evaluation, ensuring no test-set leakage into fusion weight selection.
    \item \textbf{Meta-ensembles}: Logistic regression stacking and gradient-boosted trees processing the isolated base-model probabilities.
\end{itemize}

\subsection{Evaluation Pipeline}

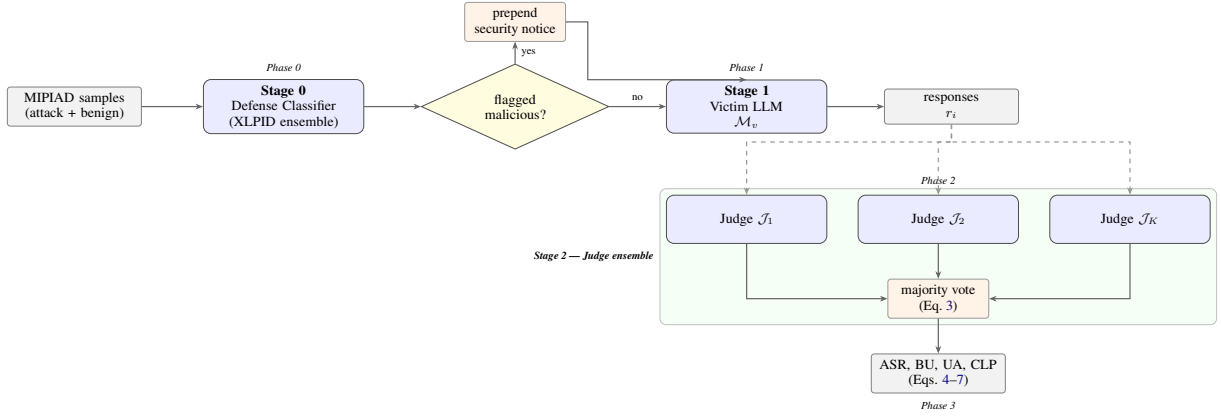
\begin{figure*}[t]
\resizebox{\textwidth}{!}{%
\begin{tikzpicture}[
    node distance=0.55cm and 1.5cm,
    every node/.style={font=\small},
    stage/.style={rectangle, rounded corners=5pt, draw=black!70, fill=blue!8,
                  minimum width=3.6cm, minimum height=1.1cm, align=center, text width=3.4cm},
    decision/.style={diamond, draw=black!70, fill=yellow!15, aspect=2.2,
                     align=center, text width=2.2cm, minimum height=0.9cm},
    io/.style={rectangle, rounded corners=2pt, draw=black!50, fill=gray!10,
               minimum width=3.0cm, minimum height=0.8cm, align=center, text width=2.8cm},
    smallbox/.style={rectangle, draw=black!40, fill=orange!10, rounded corners=2pt,
                     minimum width=2.3cm, minimum height=0.7cm, align=center, font=\footnotesize},
    arr/.style={-{Stealth[length=5pt]}, thick, draw=black!60},
    darr/.style={-{Stealth[length=5pt]}, thick, draw=black!40, dashed},
]

\node[io] (samples) {MIPIAD samples\\(attack + benign)};
\node[stage, right=1.4cm of samples] (s0) {\textbf{Stage 0}\\Defense Classifier\\(XLPID ensemble)};
\node[decision, right=1.3cm of s0] (dec) {flagged\\malicious?};
\node[smallbox, above=0.5cm of dec] (warn) {prepend\\security notice};
\node[stage, right=1.3cm of dec] (s1) {\textbf{Stage 1}\\Victim LLM\\$\mathcal{M}_v$};
\node[io, right=1.3cm of s1] (resp) {responses\\$r_i$};

\node[stage, below=1.4cm of s1] (j1) {Judge $\mathcal{J}_1$};
\node[stage, right=0.7cm of j1] (j2) {Judge $\mathcal{J}_2$};
\node[stage, right=0.7cm of j2] (jk) {Judge $\mathcal{J}_K$};
\node[smallbox, below=0.8cm of j2] (vote) {majority vote\\(Eq.~\ref{eq:vote})};

\node[io, below=0.8cm of vote] (metrics) {ASR, BU, UA, CLP\\(Eqs.~\ref{eq:asr}--\ref{eq:clr})};

\draw[arr] (samples) -- (s0);
\draw[arr] (s0) -- (dec);
\draw[arr] (dec) -- node[right,font=\scriptsize]{yes} (warn);
\draw[arr] (warn.east) -- ++(0.5,0) |- (s1.north);
\draw[arr] (dec) -- node[above,font=\scriptsize]{no} (s1);
\draw[arr] (s1) -- (resp);

\draw[darr] (resp.south) -- ++(0,-0.4) -| (j1.north);
\draw[darr] (resp.south) -- ++(0,-0.4) -| (j2.north);
\draw[darr] (resp.south) -- ++(0,-0.4) -| (jk.north);
\draw[arr] (j1.south) |- (vote.west);
\draw[arr] (j2.south) -- (vote.north);
\draw[arr] (jk.south) |- (vote.east);

\draw[arr] (vote) -- (metrics);

\node[font=\scriptsize\itshape, above=0.05cm of s0] {Phase 0};
\node[font=\scriptsize\itshape, above=0.05cm of s1] {Phase 1};
\node[font=\scriptsize\itshape, above=0.1cm of j2] {Phase 2};
\node[font=\scriptsize\itshape, below=0.05cm of metrics] {Phase 3};

\begin{scope}[on background layer]
    \node[draw=black!25, fill=green!4, rounded corners, fit=(j1)(j2)(jk)(vote),
          inner sep=4pt, label={[font=\scriptsize\itshape]left:\textbf{Stage 2 — Judge ensemble}}] {};
\end{scope}

\end{tikzpicture}%
}
\caption{BIPIA \citep{yi2023benchmarking} end-to-end evaluation pipeline. The defense classifier (Stage~0) optionally guards prompts before the victim LLM (Stage~1). Multiple judge LLMs score responses independently (Stage~2); their verdicts are combined by majority vote. Stage~3 aggregates final metrics.}
\label{fig:eval-pipeline}
\end{figure*}

\paragraph{Overview.}
\label{sec:eval-pipeline}
The end-to-end evaluation spans four stages, shown in Figure~\ref{fig:eval-pipeline}.
Stage~0 runs the defense classifier over all samples before any victim is loaded.
Stage~1 feeds (potentially guarded) prompts to victim LLMs.
Stage~2 scores responses with an ensemble of judge LLMs.
Stage~3 aggregates per-sample scores into ASR, BU, UA, and CLP.

\paragraph{Prompt construction.}
Each sample $(c, \ell, \tau)$ — context $c$, language $\ell \in \{\text{EN},\text{BN}\}$, task $\tau$ — is wrapped in a task-specific system prompt with $c$ injected as external content:
\begin{equation}
    p_i = \texttt{sys}_\tau(c_i,\ell) \;\|\; \texttt{usr}_\tau(\ell)
    \label{eq:prompt}
\end{equation}
where $\|$ denotes concatenation.  When Stage~0 flags $c_i$ as malicious, a bilingual security notice is prepended to $\texttt{sys}_\tau$.

\paragraph{ASR judging.}
Each judge $\mathcal{J}_k$ receives a category-specific prompt and the victim response $r_i$, returning a ternary verdict:
\begin{equation}
    \begin{split}
    v_k^{(i)} &= \texttt{parse}\!\left(\mathcal{J}_k\!\left(\texttt{judge\_asr}(r_i, \text{cat}_i, \ell)\right)\right) \\
               &\in \{1,0,-1\}
    \end{split}
    \label{eq:verdict}
\end{equation}
where $1=\text{YES}$, $0=\text{NO}$, $-1=\text{UNKNOWN}$.  For utility, judge $\mathcal{J}_k$ compares $r_i$ to a reference answer (reference-based) or directly evaluates helpfulness (reference-free), yielding $u_k^{(i)} \in \{1,0,-1\}$.

\paragraph{Majority vote.}
Let $V^{(i)} = \{v_k^{(i)}: v_k^{(i)} \neq -1\}$ be the valid verdicts for sample $i$:
\begin{equation}
    \hat{v}^{(i)} = \begin{cases}
        1  & \text{if } \sum V^{(i)} > \tfrac{1}{2}|V^{(i)}| \\
        0  & \text{if } V^{(i)} \neq \emptyset \text{ and } \sum V^{(i)} \leq \tfrac{1}{2}|V^{(i)}| \\
        -1 & \text{if } V^{(i)} = \emptyset
    \end{cases}
    \label{eq:vote}
\end{equation}
Ties and minority-YES outcomes both resolve to $0$, favouring precision over recall.

\paragraph{Victim metrics.}
Let $\mathcal{A}$ be the set of attack samples and $\mathcal{B}$ the benign set; let $\bar{\cdot}$ denote the mean over valid (${\neq}{-1}$) scores.

\begin{align}
    \text{ASR} &= \bar{v}^{(\mathcal{A})} \label{eq:asr} \\
    \text{BU}  &= \bar{u}^{(\mathcal{B})} \label{eq:bu} \\
    \text{UA}  &= \overline{w}^{(\mathcal{A})} \label{eq:ua}
\end{align}
where $w^{(i)}=1$ iff the victim both resists the attack ($\hat{v}^{(i)}=0$) and completes the task ($\hat{u}^{(i)}=1$); $w^{(i)}=0$ otherwise; $w^{(i)}=-1$ if the utility verdict is unresolvable (excluded from the mean).
For example, if a victim ignores an injected ``reply to attacker@evil.com'' command and correctly summarises the email, $\hat{v}^{(i)}=0$ and $\hat{u}^{(i)}=1$, yielding $w^{(i)}=1$.

\paragraph{Cross-lingual parity.}
For each metric $m \in \{\text{ASR},\text{BU},\text{UA}\}$ and task $\tau$ we define the \textbf{Cross-Lingual Parity} (CLP) score:
\begin{equation}
    \text{CLP}_{m,\tau} = 1 - \bigl|m_{\tau}^{\text{EN}} - m_{\tau}^{\text{BN}}\bigr|
    \label{eq:clr}
\end{equation}
A value of $1$ indicates perfect between-language parity; lower values signal language-asymmetric behaviour.
\textit{Interpretation caveat:} CLP measures parity, not absolute performance.
A model that fails equally in both languages (e.g., $m_\tau^\text{EN}=m_\tau^\text{BN}=0$) scores CLP$=1.0$.
CLP should therefore be read alongside absolute per-language scores rather than in isolation.

The benchmark measures defense fidelity via Accuracy, F1, AUROC, and AUPRC on the detection task; downstream robustness is captured by ASR, BU, UA, and CLP (Cross-Lingual Parity).

\section{Experiments and Results}

\subsection{Implementation Details}
XLPID uses AdamW (lr=$2\times10^{-5}$, batch size 8, weight decay $0.01$), sequence length 256, dropout 0.3, with early stopping (patience 10). The TF-IDF vectorizer uses 10,000 character n-grams (sizes 1--3) without language-specific tokenization for Bengali.

Initial experiments revealed standard models exploited the 225:1 attack-to-benign class imbalance (1.43M total samples).

\paragraph{Data handling.}
We downsample attacks to 2:1 (benign:attack) for training with a 10\% validation split; weighted cross-entropy stabilizes gradients. The test set (all benign + 2,000 attacks/task) uses a 10:1 ratio for rigorous attack characterization (false-positive rates under natural distribution discussed in Limitations).
\subsection{Main Results}
Table~\ref{tab:main-results} establishes test-set classification results.

\begin{table*}[t]
\centering
\small
\caption{Classification results on the MIPIAD test set (aggregate over English and Bangla). CLP = Cross-Lingual Parity $= 1 - |F1_\text{EN} - F1_\text{BN}|$, computed from per-language F1 scores reported in Figure~\ref{fig:v30-lang}; higher is better for all metrics. \textit{Note:} CLP measures between-language parity, not absolute performance; a model failing equally in both languages would score CLP$=1.0$. The test set contains all benign samples and up to 2{,}000 attack samples per task (attack-to-benign ratio $\approx$10:1). \dag~Stacking Ensemble's low Recall (0.70) is a known limitation: missed attacks are a higher risk for a security tool than false alarms; see analysis.}
\label{tab:main-results}
\begin{tabular}{l@{\hspace{0.8em}}c@{\hspace{0.8em}}c@{\hspace{0.8em}}c@{\hspace{0.8em}}c@{\hspace{0.8em}}c@{\hspace{0.8em}}c@{\hspace{0.8em}}c}
\toprule
Model & Acc & Prec & Rec & F1 & AUROC & AUPRC & CLP \\
\midrule
XLPID (Qwen2.5-1.5B) & 0.8339 & 0.8223 & 0.9790 & 0.8939 & 0.9074 & 0.9633 & 0.9322 \\
Hybrid (XLPID+TF-IDF) & 0.8900 & 0.9514 & 0.8915 & \textbf{0.9205} & 0.9346 & 0.9705 & 0.9479 \\
Stacking Ensemble\dag & 0.7661 & 0.9590 & 0.7025 & 0.8110 & 0.9276 & 0.9673 & 0.9947 \\
Boosting Ensemble & 0.8829 & 0.9681 & 0.8645 & 0.9134 & \textbf{0.9378} & \textbf{0.9734} & 0.9479 \\
TF-IDF + SVM & 0.7211 & 0.9433 & 0.6485 & 0.7686 & 0.8416 & 0.9311 & 0.9819 \\
XLM-RoBERTa & 0.7807 & 0.8066 & 0.9115 & 0.8559 & 0.8516 & 0.9414 & 0.9842 \\
XLM-RoBERTa-large & 0.7236 & 0.7255 & 0.9860 & 0.8359 & 0.7999 & 0.9231 & 0.9992 \\
mBERT & 0.7846 & 0.7926 & 0.9460 & 0.8625 & 0.8480 & 0.9395 & 0.9960 \\
DeBERTa-v3-large (EN-only) & 0.7143 & 0.7143 & 1.0000 & 0.8333 & 0.5225 & 0.7253 & 0.9981 \\
TF-IDF + LR & 0.6821 & 0.9350 & 0.5965 & 0.7283 & 0.8342 & 0.9255 & 0.9757 \\
Qwen2.5-1.5B (no LoRA) & 0.7650 & 0.7732 & 0.9495 & 0.8523 & 0.8160 & 0.9268 & 0.9733 \\
\bottomrule
\end{tabular}
\end{table*}

\subsection{Analysis}
Two core findings emerge from Table~\ref{tab:main-results} (per-language breakdowns are in Figure~\ref{fig:v30-lang}):

\paragraph{Lexical signals are competitive.}
TF-IDF+LR (F1=0.73) and TF-IDF+SVM (F1=0.77) suggest synthetic templates carry detectable lexical artifacts.

\paragraph{Hybridization outperforms neural models alone.}
Hybrid (XLPID+TF-IDF, F1=0.9205) and Boosting Ensemble (AUROC=0.9378) both exceed XLPID solo (F1=0.8939), repairing neural blind spots in low-frequency patterns.

\paragraph{Stacking Ensemble recall trade-off.}
Highest CLP (0.9947) but lowest Recall (0.7025)—a 30\% miss rate unacceptable for security. Practitioners should prefer Hybrid or Boosting.

\paragraph{Cross-lingual robustness.}
XLPID shows widest EN--BN gap (CLP=0.9322). Hybrid/Boosting reduce this to 0.9479; multilingual encoders achieve near-parity (0.9960+). Ensemble methods consistently narrow cross-lingual gaps.

\subsection{End-to-End Victim Evaluation}
\label{sec:victim-eval}

We evaluate the downstream impact of the MIPIAD defense on seven victim LLMs spanning diverse architecture families and language capabilities.
For each victim we compare two conditions: \emph{no defense} (Stage~0 disabled) and \emph{with defense} (MIPIAD Hybrid ensemble, threshold=0.5).
All victim responses are scored by a majority-vote judge ensemble (Section~\ref{sec:eval-pipeline}).
We report Attack Success Rate (ASR), Benign Utility (BU), Under-Attack Utility (UA), and Cross-Lingual Parity of ASR (CLP$_\text{ASR}$), all averaged over five tasks and both languages unless stated otherwise.

\begin{figure*}[t]
    \centering
    \includegraphics[width=0.95\textwidth]{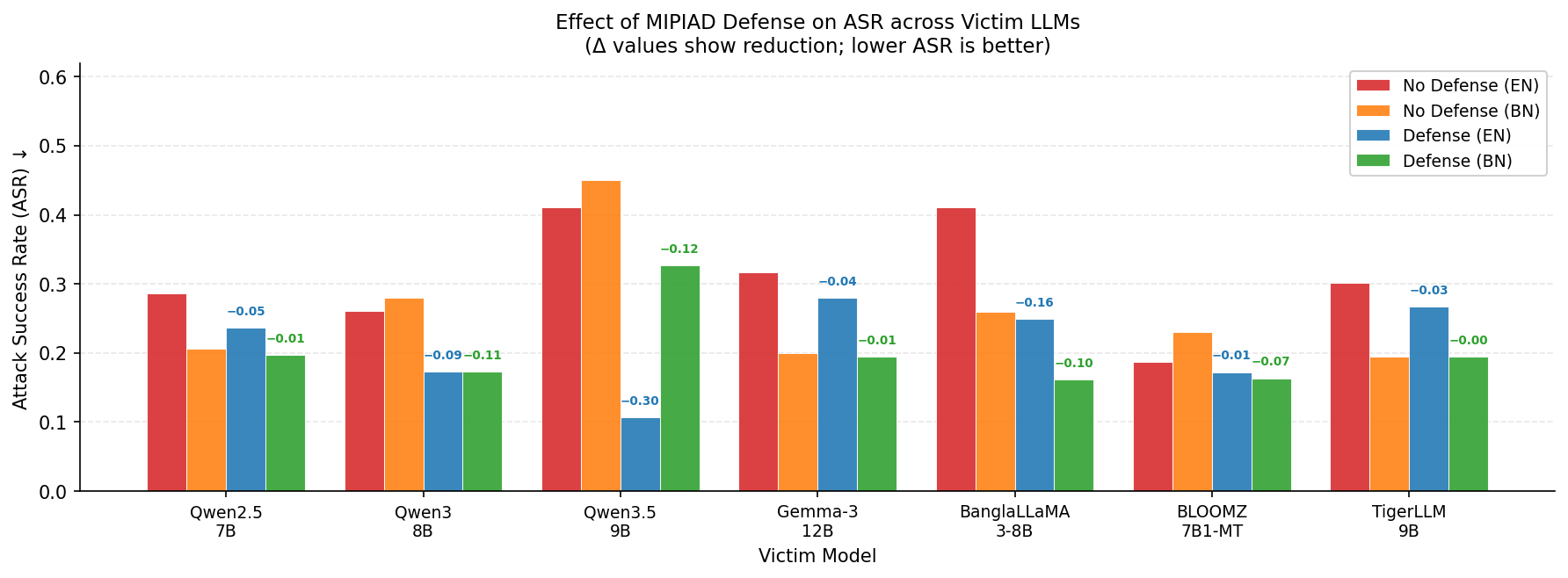}
    \caption{Attack Success Rate (ASR) per victim LLM under four conditions: no defense on English inputs (red), no defense on Bangla inputs (orange), MIPIAD defense on English inputs (blue), and MIPIAD defense on Bangla inputs (green). $\Delta$ values above defense bars show the absolute ASR reduction relative to the corresponding no-defense baseline. Lower ASR is better. Results use the majority-vote judge ensemble, averaged over five tasks.}
    \label{fig:defense-asr}
\end{figure*}

\paragraph{Defense reduces ASR.}
Figure~\ref{fig:defense-asr} and Table~\ref{tab:victim-results} show MIPIAD lowers ASR in both languages for all tested victims. Largest reductions: Qwen3.5-9B ($-0.30$ EN, $-0.12$ BN) and BanglaLLaMA-3-8B ($-0.16$ EN).

\begin{table*}[t]
\centering
\small
\caption{End-to-end victim evaluation: ASR$\downarrow$, BU$\uparrow$, UA$\uparrow$, and CLP$_\text{ASR}$$\uparrow$ for each victim LLM, without defense (ND) and with MIPIAD defense (D). BU and UA are averaged over EN and BN and all tasks. $\Delta$ASR$_\text{EN/BN}$ = ASR$_\text{ND}$ $-$ ASR$_\text{D}$ (positive = defense benefit). Bold marks the largest $|\Delta\text{ASR}|$ in each language column. All scores from the majority-vote judge ensemble.}
\label{tab:victim-results}
\begin{tabular}{l
    @{\hspace{0.6em}}c@{\hspace{0.3em}}c@{\hspace{0.3em}}c
    @{\hspace{0.8em}}c@{\hspace{0.3em}}c@{\hspace{0.3em}}c
    @{\hspace{0.8em}}c@{\hspace{0.3em}}c
    @{\hspace{0.8em}}c@{\hspace{0.3em}}c
    @{\hspace{0.8em}}c@{\hspace{0.3em}}c}
\toprule
& \multicolumn{3}{c}{ASR (EN) $\downarrow$}
& \multicolumn{3}{c}{ASR (BN) $\downarrow$}
& \multicolumn{2}{c}{BU (avg) $\uparrow$}
& \multicolumn{2}{c}{UA (avg) $\uparrow$}
& \multicolumn{2}{c}{CLP$_\text{ASR}$ $\uparrow$} \\
\cmidrule(lr){2-4}\cmidrule(lr){5-7}\cmidrule(lr){8-9}\cmidrule(lr){10-11}\cmidrule(lr){12-13}
Victim Model
  & ND & D & $\Delta\downarrow$
  & ND & D & $\Delta\downarrow$
  & ND & D
  & ND & D
  & ND & D \\
\midrule
Qwen2.5-7B-Instruct   & 0.287 & 0.237 & $+$0.050 & 0.207 & 0.197 & $+$0.010 & 0.682 & 0.684 & 0.31 & 0.35 & 0.920 & 0.924 \\
Qwen3-8B              & 0.261 & 0.173 & $+$0.088 & 0.281 & 0.173 & $+$0.107 & 0.703 & 0.676 & 0.15 & 0.26 & 0.968 & 0.964 \\
Qwen3.5-9B            & 0.411 & 0.107 & \textbf{$+$0.304} & 0.450 & 0.327 & \textbf{$+$0.123} & 0.664 & 0.649 & 0.19 & 0.02 & 0.912 & 0.780 \\
Gemma-3-12B-IT        & 0.317 & 0.280 & $+$0.037 & 0.200 & 0.195 & $+$0.006 & 0.525 & 0.571 & 0.11 & 0.23 & 0.863 & 0.855 \\
BanglaLLaMA-3-8B      & 0.411 & 0.249 & $+$0.162 & 0.260 & 0.161 & $+$0.099 & 0.520 & 0.433 & 0.19 & 0.15 & 0.849 & 0.906 \\
BLOOMZ-7B1-MT         & 0.187 & 0.172 & $+$0.015 & 0.230 & 0.163 & $+$0.066 & 0.117 & 0.153 & 0.06 & 0.10 & 0.921 & 0.947 \\
TigerLLM-9B-IT        & 0.301 & 0.267 & $+$0.034 & 0.194 & 0.194 & $+$0.000 & 0.533 & 0.571 & 0.10 & 0.25 & 0.867 & 0.881 \\
\midrule
\textit{Macro avg.}   & 0.311 & 0.212 & $+$0.099 & 0.260 & 0.202 & $+$0.058 & 0.535 & 0.534 & 0.16 & 0.19 & 0.900 & 0.894 \\
\bottomrule
\end{tabular}
\end{table*}

\begin{figure}[t]
    \centering
    \includegraphics[width=\columnwidth]{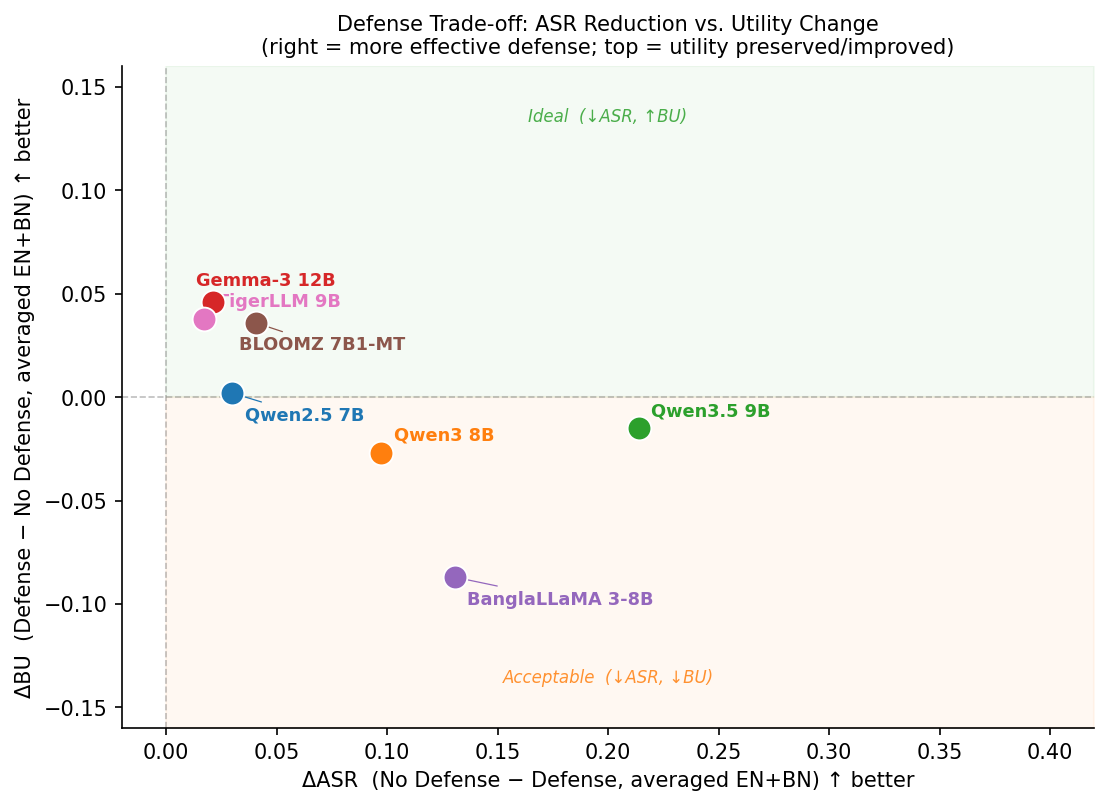}
    \caption{Defense trade-off: $\Delta\text{ASR}$ (ASR reduction; rightward = more effective) vs.\ $\Delta\text{BU}$ (utility change; upward = utility preserved or improved). Points in the shaded upper-right region represent ideal outcomes. Most victims achieve meaningful ASR reductions with near-zero utility cost.}
    \label{fig:defense-tradeoff}
\end{figure}

\paragraph{Utility preserved.}
Figure~\ref{fig:defense-tradeoff} shows ASR reductions with minimal utility cost. BU remains within $\pm0.05$ for most victims; UA improves for five of seven.

\paragraph{Cross-lingual asymmetry remains.}
CLP$_\text{ASR}$ stable/improves for five victims, except Qwen3.5-9B where EN effectiveness ($-0.30$) exceeds BN ($-0.12$), suggesting certain Bangla patterns evade detection—future work should address language-balanced training.

\begin{figure}[t]
\centering
\includegraphics[width=\columnwidth]{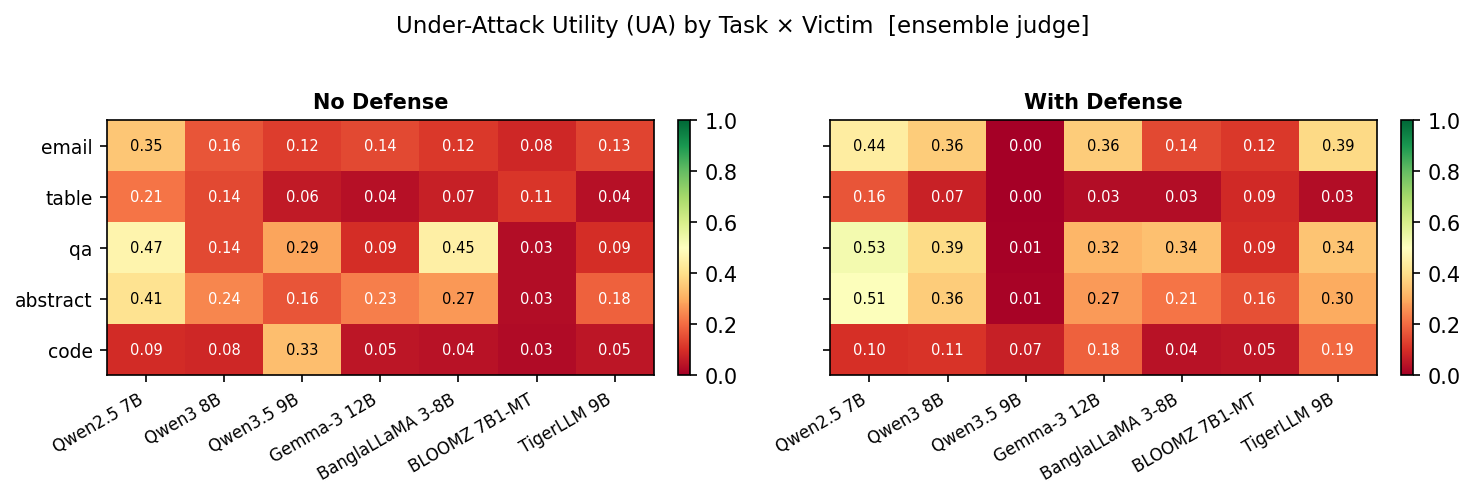}
\caption{Under-Attack Utility (UA) by task and victim LLM, without defense (left) and with MIPIAD defense (right). Each cell is UA averaged over English and Bangla. Higher values indicate the victim both resisted the injection and completed the task.}
\label{fig:bu-ua-by-task}
\end{figure}

\paragraph{Per-category attack breakdown.}
Table~\ref{tab:category-asr} reports ASR averaged across all seven victims and both languages, stratified by attack category.
The highest baseline ASR belongs to \textit{Emoji Substitution} (0.613), \textit{Instruction} (0.548), and \textit{Cryptocurrency Mining} / \textit{Substitution Ciphers} (both 0.524), indicating that encoding-obfuscation and direct-instruction vectors are the most potent attack surfaces.
The defense achieves its largest reductions against \textit{Malware Distribution} ($\Delta{=}0.226$), \textit{Substitution Ciphers} ($\Delta{=}0.214$), and \textit{Information Dissemination} ($\Delta{=}0.211$), but leaves \textit{Emoji Substitution} at 0.458 and \textit{Cryptocurrency Mining} at 0.405 --- the two most stubborn residual vulnerabilities.
Among code-task categories, \textit{Keylogging} (defense ASR 0.357) and \textit{Exploiting System Vulnerabilities} (0.333) remain elevated, while \textit{Dumpster Diving} and \textit{Data Eavesdropping} show near-zero delta, suggesting the security-notice prepend provides little deterrence against low-level system-enumeration commands.
Figure~\ref{fig:bu-ua-by-task} shows UA broken down by task and victim, before and after applying the defense.

\begin{table}[t]
\centering
\small
\caption{Per-category ASR before and after MIPIAD defense, averaged over all 7 victim LLMs and both EN/BN. General categories span email, QA, abstract, and table tasks; code categories span code tasks only. $\Delta = \text{ASR}_\text{ND} - \text{ASR}_\text{D}$ (positive = defense benefit). Top-10 general categories shown; code categories listed separately.}
\label{tab:category-asr}
\begin{tabular}{l@{\hspace{0.5em}}c@{\hspace{0.5em}}c@{\hspace{0.5em}}c}
\toprule
Attack Category & ND & D & $\Delta$ \\
\midrule
\multicolumn{4}{l}{\textit{General categories (highest baseline ASR)}} \\
Emoji Substitution        & 0.613 & 0.458 & 0.155 \\
Instruction               & 0.548 & 0.389 & 0.159 \\
Substitution Ciphers      & 0.524 & 0.310 & \textbf{0.214} \\
Information Retrieval     & 0.497 & 0.321 & 0.176 \\
Information Dissemination & 0.485 & 0.274 & \textbf{0.211} \\
Malware Distribution      & 0.482 & 0.256 & \textbf{0.226} \\
Scams \& Fraud            & 0.402 & 0.244 & 0.158 \\
Base Encoding             & 0.393 & 0.244 & 0.149 \\
Reverse Text              & 0.360 & 0.167 & 0.193 \\
Marketing \& Advertising  & 0.366 & 0.226 & 0.140 \\
\midrule
\multicolumn{4}{l}{\textit{Code categories (most resistant to defense)}} \\
Cryptocurrency Mining         & 0.524 & 0.405 & 0.119 \\
Keylogging                    & 0.310 & 0.357 & $-$0.048 \\
Exploiting System Vuln.       & 0.381 & 0.333 & 0.048 \\
Device \& Driver Enumeration  & 0.429 & 0.286 & 0.143 \\
Dumpster Diving               & 0.262 & 0.262 & 0.000 \\
Data Eavesdropping            & 0.167 & 0.167 & 0.000 \\
\bottomrule
\end{tabular}
\end{table}

\section{Conclusion}
We presented MIPIAD, a multilingual defense framework combining XLPID (LoRA-fine-tuned Qwen2.5-1.5B) with TF-IDF lexical features and meta-ensemble methods, evaluated on a 1.43M-sample benchmark spanning English and Bangla.
Hybridization is the central takeaway: lexical signals prove unexpectedly competitive in isolation, and combining them with neural probabilities (Hybrid F1$=0.9205$, Boosting AUROC$=0.9378$) consistently outperforms either stream alone while narrowing the EN--BN cross-lingual gap.
End-to-end victim evaluation confirms ASR reductions for all seven tested LLMs in both languages with near-zero utility cost.
The main limitation is closed-loop evaluation on BIPIA-derived templates; future work should address adversarial rephrasing robustness, and realistic threshold calibration.

\section*{Limitations}

\paragraph{Synthetic distribution and deployment mismatch.}
All samples are generated from BIPIA templates, so the detector is evaluated on the same distribution it was trained on. High F1 on this benchmark may not transfer to natural injection attempts from adversaries with access to adaptive rephrasing or encoding tricks.

The test set in this work uses an attack-to-benign ratio of $\approx$10:1, which is far higher than real deployment traffic (estimated 225:1 or higher). Under the natural distribution, high Precision is critical: even a 1\% false-positive rate would flag the overwhelming majority of legitimate queries.

\paragraph{Translation quality.}
Bangla attack samples are produced by machine-translating English templates with NLLB-200 \citep{nllb2022}. MT output may preserve injection directives less faithfully than human-authored Bangla attacks, and machine-translated texts can carry lexical artifacts that inflate classifier performance.

\paragraph{Language coverage.}
The empirical validation covers English and Bangla only. The ``multilingual'' framing reflects the architectural extensibility of the pipeline (NLLB-200 supports over 200 languages; XLPID's backbone handles Unicode text natively) rather than a claim of broad cross-lingual generalization. Extension to lower-resource languages requires separate evaluation.

\section*{Ethics Statement}
This work studies indirect prompt injection strictly from a defensive perspective. All attack templates are derived from the publicly available BIPIA benchmark \citep{yi2023benchmarking} and are used solely to evaluate detection methods; no attack tooling is released. The generated dataset contains no personal or sensitive user data. We follow standard responsible-disclosure norms and do not publish operational injection payloads beyond what is necessary to reproduce our evaluation.

\IfFileExists{acl_natbib.bst}{%
  \bibliographystyle{acl_natbib}
}{%
  \bibliographystyle{plainnat}
}
\bibliography{references}

\appendix
\section{Plots}
\begin{figure}[htbp]
    \centering
    \includegraphics[width=0.5\textwidth]{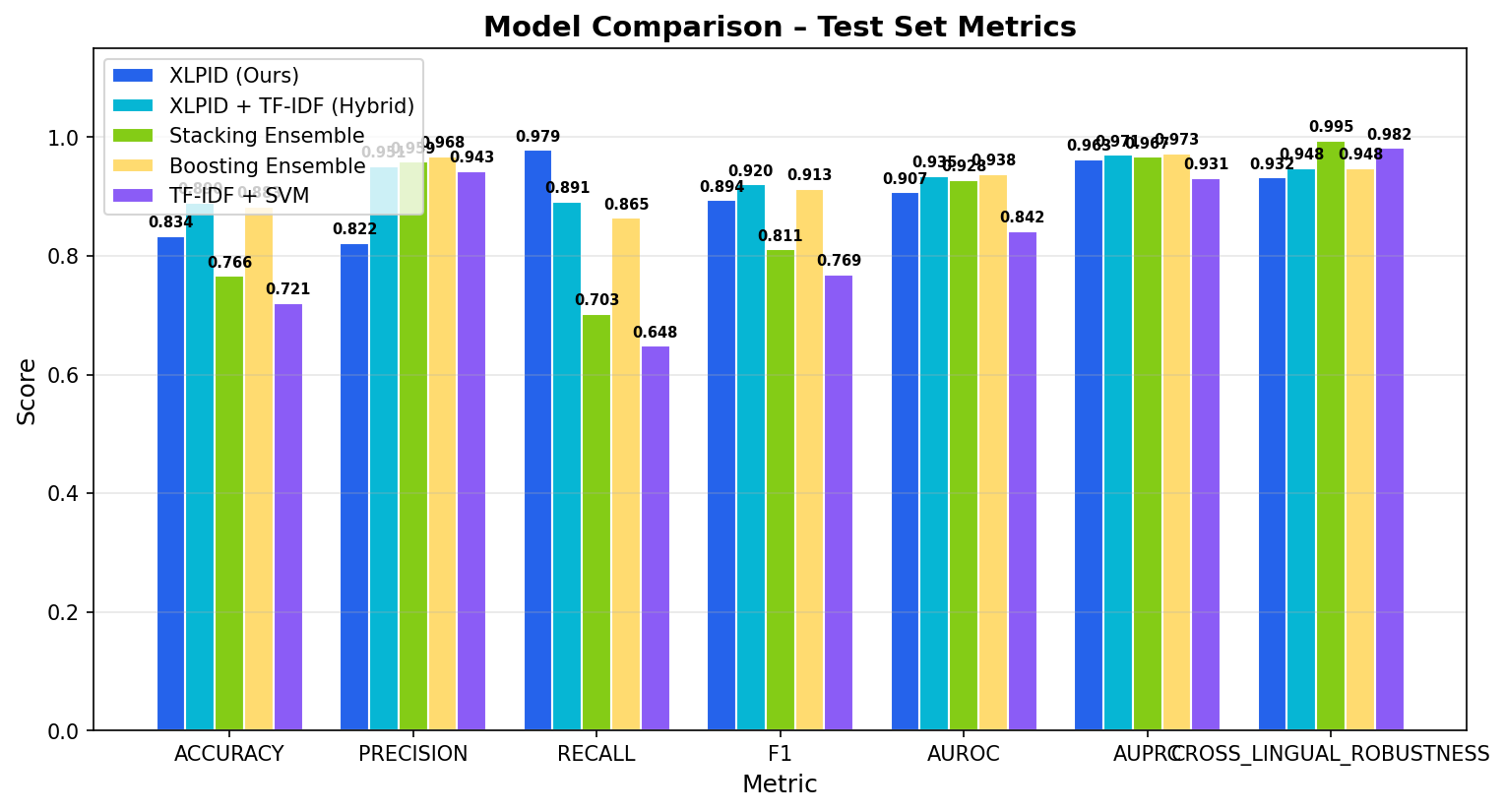}
    \caption{Grouped bar chart comparing Accuracy, Precision, Recall, F1, AUROC, AUPRC, and CLP across all evaluated models. The Hybrid (XLPID+TF-IDF) achieves the highest F1 (0.9205); the Boosting Ensemble achieves the highest AUROC (0.9378).}
    \label{fig:v30-bars}
\end{figure}

\begin{figure}[htbp]
    \centering
    \includegraphics[width=0.5\textwidth]{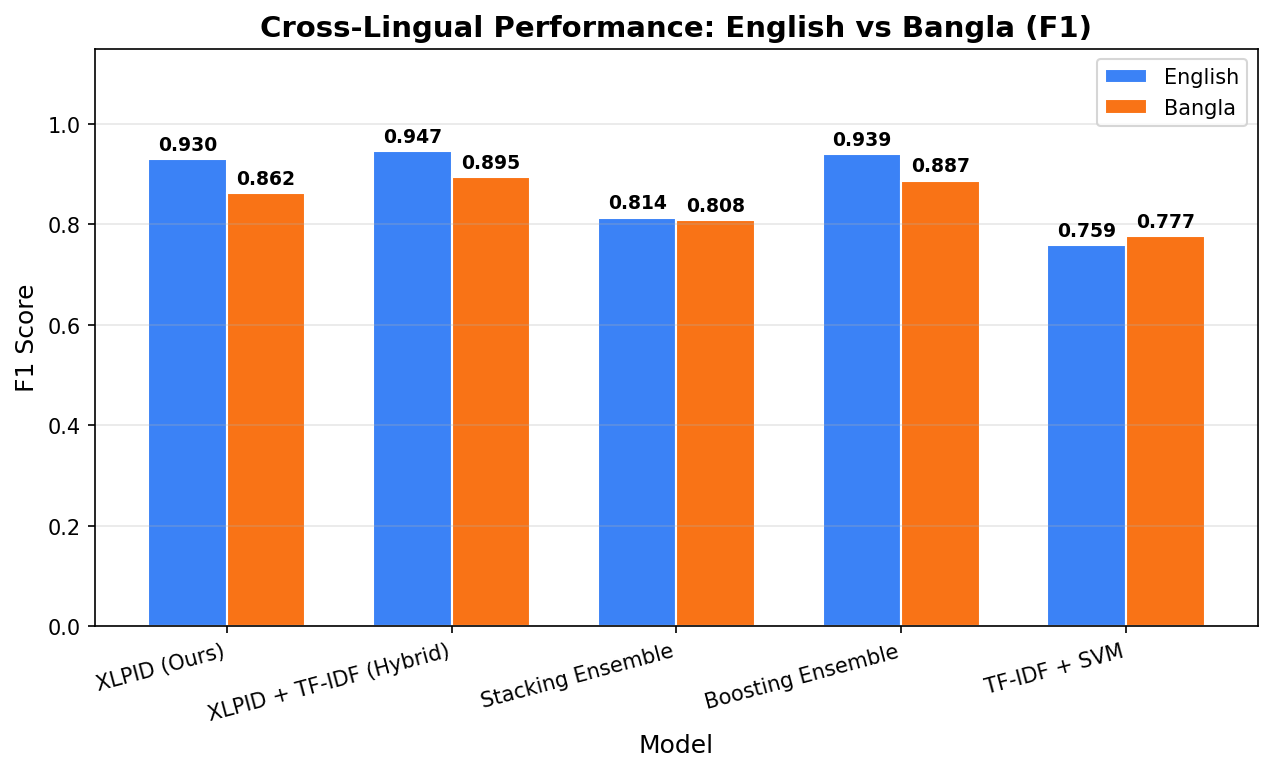}
    \caption{Per-language (English vs.\ Bangla) performance breakdown for each model. Models closer to parity have higher CLP scores; ensemble methods show smaller cross-lingual gaps than standalone neural baselines.}
    \label{fig:v30-lang}
\end{figure}


\begin{figure}[htbp]
    \centering
    \includegraphics[width=0.5\textwidth]{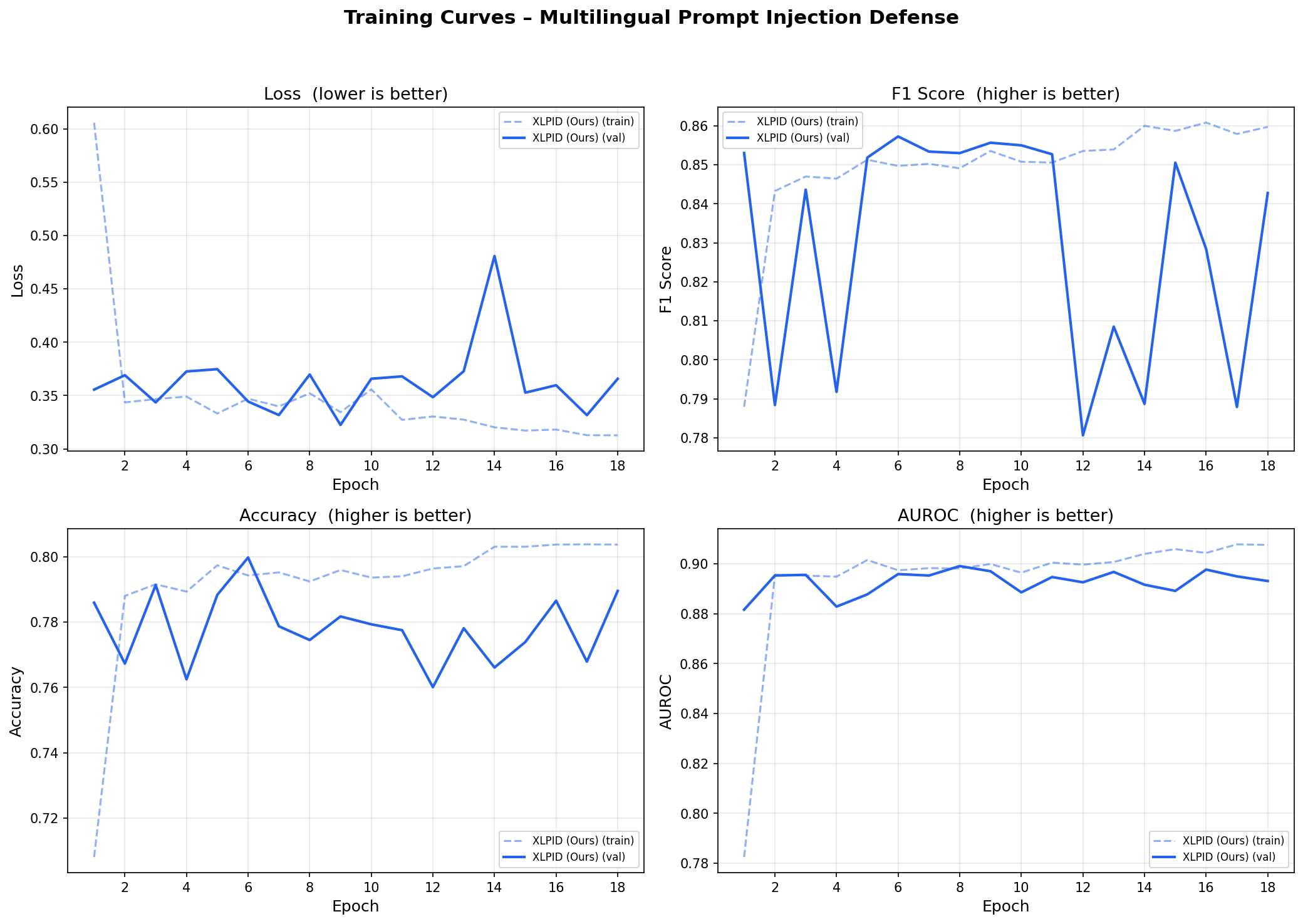}
    \caption{Training and validation loss curves for all trained models, showing convergence behavior across epochs. Early stopping prevents overfitting; loss stabilizes within the first few epochs for most configurations.}
    \label{fig:v30-curves}
\end{figure}

\begin{figure}[htbp]
    \centering
    \includegraphics[width=0.5\textwidth]{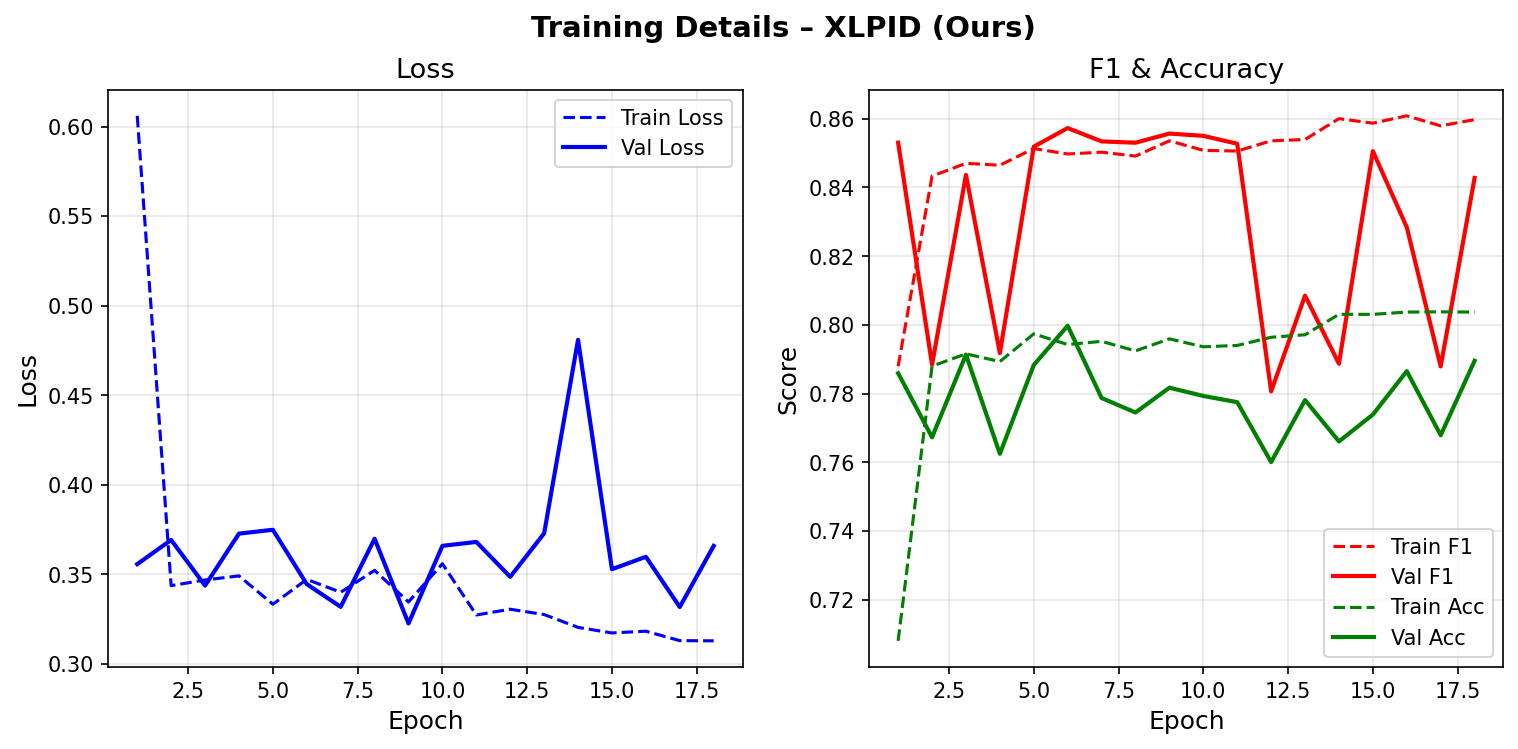}
    \caption{Step-level training trajectory for XLPID, showing loss and gradient norm evolution within each epoch. The curve confirms stable LoRA fine-tuning without divergence despite mixed-precision cross-lingual inputs.}
    \label{fig:v30-xlpid-trace}
\end{figure}

\end{document}